\documentclass[runningheads]{llncs}

\usepackage{eccv}

\usepackage{eccvabbrv}

\usepackage{graphicx}
\usepackage{booktabs}
\usepackage{multirow}
\usepackage{subcaption} % for subtable
\usepackage{makecell}

\usepackage[accsupp]{axessibility}  % Improves PDF readability for those with disabilities.
\usepackage{mathtools}
\newcommand{\defeq}{\vcentcolon=}

\usepackage[dvipsnames]{xcolor}
\usepackage{booktabs}

\newcommand{\hod}{\texttt{HO3D} }
\newcommand{\eoat}{\texttt{YCBInEOAT} }
\newcommand{\multitrack}{\texttt{YCBMultiTrack} }
\usepackage{hyperref}

\usepackage{orcidlink}

\begin{document}

% ---------------------------------------------------------------
% TODO REVIEW: Replace with your title
\title{Point2Pose: Occlusion-Recovering 6D Pose Tracking and 3D Reconstruction for Multiple Unknown Objects via 2D Point Trackers} 

% TODO REVIEW: If the paper title is too long for the running head, you can set
% an abbreviated paper title here. If not, comment out.
\titlerunning{Point2Pose: Occlusion-Recovering 6D Pose Tracking}

% TODO FINAL: Replace with your author list. 
% Include the authors' OCRID for the camera-ready version, if at all possible.
\author{Tzu-Yuan Lin\inst{1} \and
Ho Jae Lee\inst{1} \and
Kevin Doherty\inst{2}\textsuperscript{\S} \and
Yonghyeon Lee\inst{1} \and \\ Sangbae Kim\inst{1}}

% TODO FINAL: Replace with an abbreviated list of authors.
\authorrunning{T.-Y. Lin et al.}
% First names are abbreviated in the running head.
% If there are more than two authors, 'et al.' is used.

% TODO FINAL: Replace with your institution list.
\institute{Massachusetts Institute of Technology, USA\\ 
\email{\{tzuyuan, hjlee201, yhl, sangbae\}@mit.edu}
\and Boston Dynamics, USA \\
\email{kdoherty@mit.edu}}
\maketitle

\footnotetext{$^{\S}$ Work was conducted in personal time and independently of the author's affiliated organization.}

\begin{abstract}
  We present \textit{Point2Pose}, a model-free method for causal 6D pose tracking of multiple rigid objects from monocular RGB-D video. Initialized only from sparse image points on the objects, our approach tracks multiple unseen objects without requiring object CAD models or category priors. Point2Pose leverages a 2D point tracker to obtain long-range correspondences, enabling instant recovery after complete occlusion. Simultaneously, the system incrementally reconstructs an online Truncated Signed Distance Function (TSDF) representation of the tracked targets. Alongside the method, we introduce a new multi-object tracking dataset comprising both simulation and real-world sequences, with motion-capture ground truth for evaluation. Experiments show that Point2Pose trades some single-object pose accuracy for broader model-free tracking capabilities, including multi-object tracking and recovery from complete occlusion. Project page: \url{https://point2pose.github.io/}.
  
\end{abstract}

\begin{figure}[htbp]
    \centering
    \vspace{-2mm}
    \includegraphics[width=\textwidth]{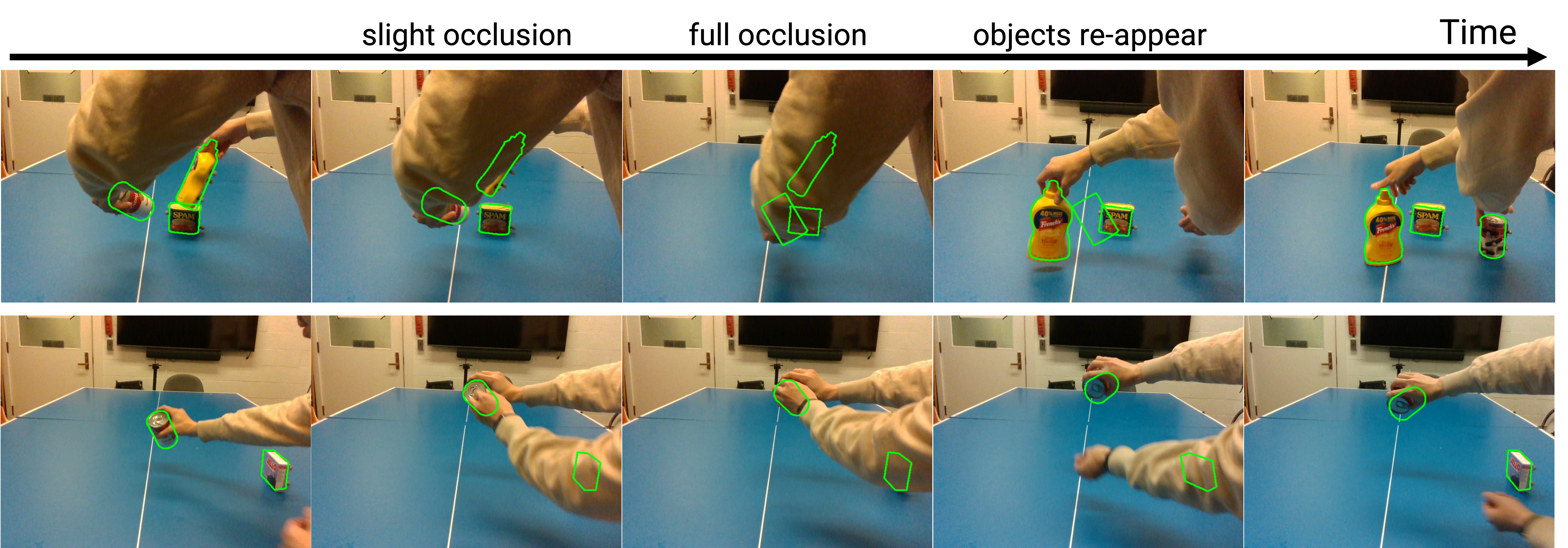}
    \caption{
    Point2Pose tracks multiple unknown objects through severe hand-object occlusion and recovers their poses when they reappear, without requiring object CAD models.}
    \label{fig:real_data}
    \vspace{-3mm}
\end{figure}
\section{Introduction}
\label{sec:intro}

Estimating the 6D pose of rigid objects from visual observations is a core problem in robotics and augmented reality. Reliable pose tracking enables agents to plan, interact with, and manipulate objects. While recent progress in object pose estimation and tracking has been substantial, many high-performing systems assume access to object CAD models~\cite{he2020pvn3d,wen2024foundationpose}, category-level priors~\cite{ponimatkin6d}, or require multi-view reconstruction before tracking~\cite{wen2024foundationpose}. These assumptions limit deployment in open-world settings, where robots must interact with previously unseen objects without pre-built geometric models. 

% Many existing model-free trackers~\cite{wen2023bundlesdf,wen2021bundletrack} are designed primarily for single-object tracking. Furthermore, they are often not robust in recovering from complete occlusions. This limitation becomes especially pronounced in egocentric environments, where multiple objects frequently undergo complete occlusion due to robot-arm motion, inter-object crossovers, or because the camera temporarily pans away from the workspace. These scenarios could lead to complete failure of the tracking methods, as they often rely on inter-frame feature matching.
Many existing model-free trackers~\cite{wen2023bundlesdf,wen2021bundletrack} are primarily designed for single-object tracking and often struggle to recover from complete occlusions. This limitation becomes particularly pronounced in egocentric settings involving multi-object manipulation, where objects may undergo full occlusion due to robot-arm motion, inter-object crossovers, or temporary camera motion away from the workspace. In such cases, tracking can fail entirely, as these methods typically rely on inter-frame feature matching and require brittle relocalization mechanisms.

To address these challenges, we present Point2Pose, a model-free method for causal 6D pose tracking of multiple unknown rigid objects from a single RGB-D video stream. Point2Pose jointly tracks object poses online while incrementally reconstructing object-centric 3D models for each tracked object. Our key idea is to use a long-range 2D point tracker as a persistent data-association module, which eliminates the need for expensive and unreliable long-horizon feature matching. By lifting these tracked 2D points using depth data, we construct object-centric 3D keypoint maps and recover poses through map-based registration. In addition, this design naturally supports simultaneous multi-object tracking through aggregated object-specific query sets and enables immediate recovery after complete occlusion when objects reappear, without requiring a separate heavyweight and potentially brittle relocalization stage.

Furthermore, we introduce \texttt{YCBMultiTrack}, a new dataset for dynamic multi-object RGB-D pose tracking with both synthetic and real-world sequences. Existing benchmarks such as HO3D~\cite{hampali2020honnotate} and YCB-InEOAT~\cite{wen2020se} focus on single-object manipulation, while YCB-Video~\cite{xiang2017posecnn} mainly contains static multi-object scenes and egocentric datasets such as HOT3D~\cite{banerjee2025hot3d} rely on multi-view RGB streams without dense depth. In contrast, \texttt{YCBMultiTrack} captures dynamic multi-object motion with inter-object occlusions and provides ground-truth object poses from simulation or motion capture. In summary, our main contributions include:
\begin{itemize}
\item We propose \textbf{Point2Pose}, a model-free method for 6D pose tracking of multiple unknown rigid objects from RGB-D video, using long-range 2D point tracks for persistent data association and recovery after complete occlusion.
\item We introduce \texttt{YCBMultiTrack}, a new dataset for dynamic multi-object RGB-D pose tracking with synthetic and real-world sequences.
\item Experiments on public benchmarks and \texttt{YCBMultiTrack} show that Point2Pose extends model-free 6D tracking to challenging multi-object scenarios with complete occlusion and object re-entry.
\item Code and dataset will be released on the project page.
\end{itemize}

% We address model-free 6D pose tracking of multiple rigid objects from monocular RGB-D video. Initialized from sparse user-provided points, the proposed method tracks object-centric poses in $SE(3)$ in a fully causal setting, without relying on object CAD models or category-level priors. By maintaining persistent point trajectories over time, the system enables reliable pose recovery after complete occlusion. Furthermore, as a byproduct of tracking, the proposed method incrementally reconstructs an object-centric 3D textured mesh.

\section{Related Work}

% In this work, we focus on the model-free pose tracking problem for unknown objects. Existing approaches can be broadly categorized into two types: methods that directly estimate pose from frame-to-frame observations without maintaining a persistent geometric representation and methods that simultaneously reconstruct an object map during tracking.

% TODO is to consolidate this
\textbf{6D object pose estimation and tracking.} 6-DoF object pose estimation and tracking infer the 3D position and orientation of a target object, expressed relative to a chosen reference frame (e.g., the CAD canonical frame, the object pose in the first frame, or the camera frame). Pose estimation compares the current observation to a reference model and estimates the object’s pose in the current frame. It typically relies on a reference model, either an existing CAD model, pre-trained on selected objects~\cite{he2020pvn3d}, or one reconstructed from multi-view images~\cite{wen2024foundationpose}. When performed in real time, pose estimation operates independently at each frame, avoiding drift accumulation and enabling natural recovery from abrupt motion or tracking failure. However, it does not enforce temporal consistency and may result in frame-to-frame jitter. In contrast, pose tracking leverages temporal coherence to propagate and refine pose estimates over time, maintaining a dynamically consistent estimate of the object’s motion~\cite{deng2021poserbpf}. It can be formulated as either model-based or model-free approaches~\cite{wen2021bundletrack}. By exploiting motion continuity, pose tracking achieves smooth and computationally efficient updates, but at the risk of drift accumulation and reduced robustness under prolonged occlusions. Several methods, e.g., MegaPose \cite{labbe2023megapose}, and FoundationPose \cite{wen2024foundationpose}, can perform both global pose estimation and tracking using separate coarse estimation and local refinement steps. These methods can use the (faster) local refinement step to perform frame-to-frame tracking given an initial estimate. However, these methods struggle to regain tracking after occlusion, and data association is a challenge in the multi-object setting. CosyPose \cite{labbe2020cosypose} and KMOPS \cite{wu2026kmops} are able to provide pose estimates for multiple static objects in a scene, but do not consider the problem of persistent temporal tracking. In contrast, our method is designed to track multiple objects that are dynamically moving, maintaining precise 6D trajectories even through complex inter-object interactions and temporary disappearances.

\textbf{Simultaneous tracking and reconstruction.} When an object CAD model is not available \emph{a priori}, tracking methods must infer object pose from online observations, often while building or updating an object representation. BundleTrack \cite{wen2021bundletrack} pioneered this direction by using a graph-based optimization that segments the object and tracks it via temporal feature matching. However, its reliance on short-term frame-to-frame correspondences makes it susceptible to drift and failure during total occlusions. BundleSDF \cite{wen2023bundlesdf} improves upon this by using a neural implicit field to represent the object, enabling joint optimization of pose and shape. However, its neural-field optimization is computationally expensive, making multi-object tracking difficult, and recovery after prolonged complete occlusion remains challenging.
Recent advancements have explored alternative representations, such as 6DOPE-GS \cite{jin20256dope}, which leverages 3D Gaussian Splatting for real-time tracking and reconstruction. While Gaussian Splatting offers impressive rendering speeds and detail, these methods still struggle with re-localization after a target object completely leaves and re-enters the camera view.
In this work, we bridge these gaps by utilizing a long-range 2D point tracker as the primary data-association engine. Unlike previous methods that rely on local descriptors or dense volumetric updates, our approach uses persistent point queries to maintain identity across long temporal gaps. This allows our method to reconstruct object-centric models for multiple items simultaneously and achieve immediate re-localization after complete occlusion.

\textbf{Object tracking datasets.} 
\hod ~\cite{hampali2020honnotate} and \eoat~\cite{wen2020se} are well-established for evaluating pose tracking methods but are limited to single-object manipulation scenarios. 
Datasets that incorporate multiple objects, such as \texttt{YCB-Video}~\cite{xiang2017posecnn}, predominantly feature static scenes where only the camera is in motion, failing to capture the complexities of independent object trajectories and inter-object occlusions. Recent large-scale egocentric benchmarks like \texttt{HOT3D}~\cite{banerjee2025hot3d} capture multi-object interactions but rely entirely on multi-view RGB and monochrome streams, lacking the dense depth information. To address this gap, \multitrack is designed specifically for multi-object RGB-D pose tracking under dynamic motion and occasional complete occlusions, and includes both synthetic and real sequences with motion-capture object pose annotations.

\section{Methodology}
Our goal is to track the 6D poses of multiple rigid objects and simultaneously reconstruct their 3D meshes from a stream of monocular RGB-D images. 
We assume a fixed camera frame $C$ rigidly attached to the world and denote an RGB-D video by $\{\mathcal{F}_t\}_{t=0}^T$, where $\mathcal{F}_t = (I_t, D_t)$ contains an RGB image $I_t \in \mathbb{R}^{3\times H\times W}$ and a depth image $D_t \in \mathbb{R}^{H \times W}$. 
The rigid transformation that maps 3D points from frame $A$ to frame $B$ is represented by $T^{B}_{A} \in \mathrm{SE}(3)$, which can also be interpreted as the pose of frame $A$ expressed in frame $B$.

For each object $i$, we rigidly attach a frame $O_i$ to the object, which we refer to as the \textit{object frame}.
We denote the pose of object frame at time $t$ expressed in $C$ as $T^C_{O_{i, t}}$, and we initialize it to coincide with the camera frame $C$, i.e., $T^C_{O_{i, 0}}=I_{4\times 4}$. 
Then, multi-object tracking reduces to estimating $T^C_{O_{i, t}} = T^C_{O_{i, 0}} T^{O_{i, 0}}_{O_{i, t}}$.
% Given an object's canonical mesh frame $M_i$ defined in the mesh model, once the transformation $T_{M_i}^{O_i}$ -- which is constant over time since both frames are rigidly attached to the object -- is identified, the pose of the canonical mesh frame at time $t$ can be recovered as $T^{C}_{M_i, t} = T^C_{O_i, t} T_{M_i}^{O_i}$.

\begin{figure}[t]
    \centering
    \includegraphics[width=\linewidth]{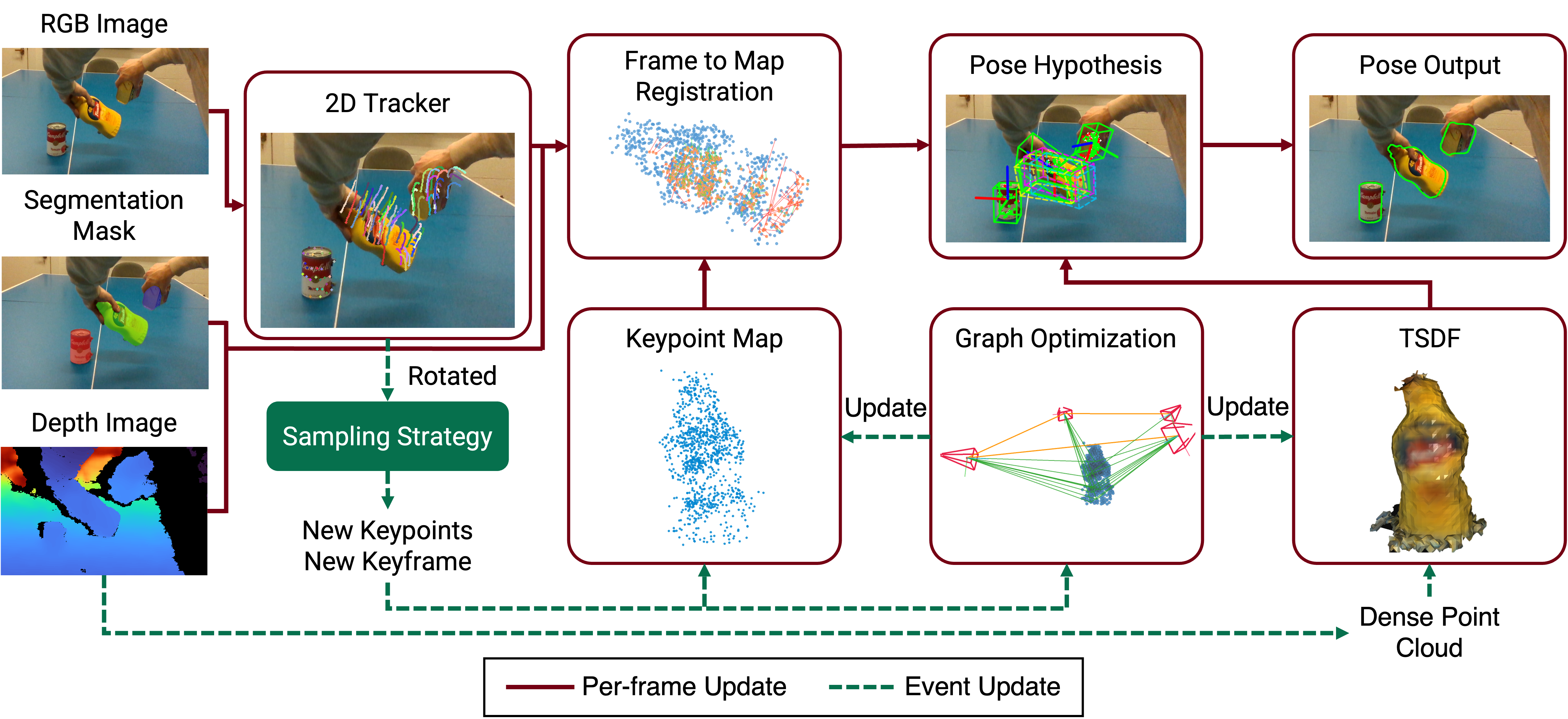}
    \caption{Overview of the Point2Pose framework. Starting from RGB-D frames and object masks, sparse object keypoints are tracked over time and used to register each frame to an object-centric map. The segmented depth observations are fused into TSDF maps for reconstruction and pose refinement, while pose-graph optimization maintains consistent trajectories across keyframes.}
    \label{fig:framework}
\end{figure}

An overview of the proposed method is shown in Fig.~\ref{fig:framework}. 
Given a few user-provided points on the objects, segmentation masks for each object are obtained using Segment Anything 2~\cite{ravi2024sam}. 
Given these masks, we sample object-specific 2D points and track them over time using a long-range point tracker, such as Track Any Points~\cite{doersch2023tapir, doersch2024bootstap} or CoTracker~\cite{karaev23cotracker, karaev24cotracker3}. 
The tracked points are lifted to 3D using depth measurements and transformed into the object frame to form an object-centric keypoint map, which we simply refer to as the map.
As the object rotates and new surfaces become visible, additional points are sampled and added to expand the map over time. Frames where new points are sampled are referred to as keyframes.
Using the natural correspondences from tracking, object poses are estimated by solving a registration problem between the current keypoint observations and the map, which we refer to as \textit{frame-to-map registration}.
Meanwhile, we perform online graph optimization to improve temporal stability and update a Truncated Signed Distance Function (TSDF) representation of the mesh model online, which is used to refine pose estimation among the pose hypotheses.
% while a keyframe graph maintains geometrically consistent object observations across time to improve temporal stability. 
% In the meantime, using the estimated object poses and the masked RGB-D point clouds, a Truncated Signed Distance Function (TSDF) are constructed online to facilitate the optimization process. 

% Figure ~\tyl{add figure} shows an overview of the proposed method. Segmentation masks of each object are obtained via Segment Anything 2~\cite{ravi2024sam}. Subsequently, 2D points are sampled within the segmented masks, and tracked by a 2D point tracker such as Track Any Points~\cite{doersch2023tapir, doersch2024bootstap} or CoTracker~\cite{karaev23cotracker, karaev24cotracker3}. 

\subsection{2D Point Tracker}
Given an RGB video $\{I_t\}_{t=0}^T$ and a set of query points, where each query point is defined as $q=(t_q, u_q)$ with $t_q$ the query frame index (i.e., the frame in which the point is queried) and $u_q \in \mathbb{R}^2$ the query pixel location in $I_{t_q}$, a 2D point tracker estimates the corresponding pixel location in $I_{t}$, denoted $u_{t} \in \mathbb{R}^2$. In addition, the tracker outputs a binary visibility indicator $\delta_{t}\in\{0,1\}$ and an uncertainty score $\sigma_{t}\in[0,1]$ for each prediction. 

In our multi-object setting, we maintain an object-specific query set $\mathcal{Q}_i$ for each object $i$, while tracking all query points jointly in a single tracker pass. Therefore, the tracker serves as a long-range data-association module, producing temporally consistent correspondences over long time horizons, bypassing the need for explicit feature matching between consecutive frames. This property enables instant recovery when the object reappears after complete occlusion. Since tracking is performed at the point level without object-specific setup, the same mechanism naturally supports multi-object tracking by operating on the aggregated query set. 

\subsection{Sampling Strategy and Keypoints Map}
When the initially tracked points become occluded or rotate out of view, additional points must be sampled to maintain tracking. These points should be chosen to jointly maximize trackability and spatial diversity to reduce the risk of degenerate pose estimation. 
To this end, we detect candidate keypoints using SuperPoint~\cite{detone2018superpoint} within the segmented mask, and use the detector confidence score as a proxy for trackability. Given a set of candidate points $\{(p_k, s_k)\}_{k=1}^{N_k}$, where $p_k$ is the pixel location and $s_k \in \left[0,1\right]$ is the confidence score from SuperPoint, we iteratively select $K \leq N_k$ points using a greedy approach. 
Let ${\cal K}$ denote the set of all keypoints already stored in the map. At each step, we choose the candidate that maximizes the following objective function, which balances trackability and spatial spread:
\begin{equation}
    J_k = \lambda s_k + (1-\lambda) \min(\frac{d_k}{r_{ideal}},1) - \beta \rho(d_k),
\end{equation}
where $d_k$ is the minimum distance between the candidate point $p_k$ and the union of the previously selected point and ${\cal K}$, $r_{ideal}$ is a parameter controlling the ideal spacing radius, and $\lambda \in \left[0,1\right]$ is a weighting parameter. We additionally include a clustering cost $\rho(d_k) = \max(0,\frac{r_{min}-d_k}{r_{min}})$ that penalizes candidates that fall below a minimum acceptable separation distance $r_{min}$ weighted by $\beta$. Intuitively, the first term rewards trackability, the second term encourages the selected points to maintain an ideal spatial spread, and the final term strictly penalizes candidates that are too close to existing tracks.

The newly selected $K$ keypoints are then lifted to 3D using the camera matrix and the depth measurements. These points are subsequently transformed into the object frame using the current estimated pose. 
Rather than immediately integrating these new keypoints into the map, they are initialized in a pending state to mitigate the impact of depth noise and incorrect pose estimate.
% A pending point is promoted to a keypoint only if it satisfies a multi-frame verification process.
% Once a candidate accumulates a sufficient streak of verified observations, it is added to the keypoint map, and the corresponding frame is designated as a keyframe.
% which we describe in detail in the Appendix.
A pending point is promoted to a keypoint only if it passes a multi-frame verification process. 
Each pending point receives a score of $+1$ or $+0$ at every time step based on the new observation. 
If the accumulated score exceeds a threshold within a fixed time window, the point is promoted to a keypoint and added to the map. 
Further details are provided in the Appendix.

\subsection{Frame-to-Map Registration for Pose Estimation}
The point tracker provides the 2D pixel location of each keypoint in the current camera frame along with an estimate of its visibility. 
By filtering out non-visible keypoints and back-projecting the remaining points using depth, we obtain 3D keypoint observations in the current camera frame and establish correspondences with the keypoint map. 
This allows us to recover the pose of the object by solving a corresponded point cloud registration problem:
\begin{equation}
    T^C_{O_{i,t}} = \underset{T \in SE(3)}{\arg\min} \sum_{n=1}^{N} \left\lVert \tilde{p}_n - T p_n\right\rVert^2,
    \label{eq:corres_svd}
\end{equation}
where $\{(\tilde{p}_n, p_n)\}_{n=1}^{N}$ denotes the set of correspondences between the 3D keypoint observations $\tilde{p}_n$ in the camera frame and their corresponding points $p_n$ in the map, where $N$ is the number of keypoint observations in the current frame.
This least-squares objective can be solved analytically using a Singular Value Decomposition (SVD) approach. 

However, because 2D point tracks often contain a high ratio of outliers due to severe occlusions or repetitive local textures, solving \eqref{eq:corres_svd} directly across all putative matches can yield catastrophic pose errors. To address this, we propose a multi-hypothesis registration strategy designed to explicitly handle geometric degeneracies and perceptual aliasing.
For instance, consider a mug rotating around its vertical axis: 2D point trackers may incorrectly predict nearly static trajectories for keypoints on the symmetric cylindrical body, while only a small subset of points on the handle reflects the true rotational motion. In such cases, single-hypothesis outlier-robust methods such as RANSAC can be dominated by the more numerous but incorrect correspondences on the body. By sequentially generating a diverse set of pose hypotheses, our approach preserves the valid transformation among the candidates.

\textbf{Multi-Hypothesis Pose Generation and Selection.}
For each object frame $O$, we generate a set of pose hypotheses $\mathcal{H} = \{(T^C_{O})_h\}_{h=1}^H$ using a sequential RANSAC scheme paired with SVD. 
% Rather than terminating after identifying a single consensus set, we iteratively extract multiple rigid transformations to account for grouped tracking failures. 
In each iteration, standard RANSAC identifies the candidate transformation $T^C_{O}$ that maximizes the number of geometric inliers, after which the corresponding consensus set is removed from the pool of correspondences.
We repeat this procedure on the remaining points. 
This greedy extraction strategy ensures that $\mathcal{H}$ captures not only the dominant motion but also alternative viable poses, preventing the system from prematurely committing to a spatially clustered set of outlier tracks.

We select the best pose hypothesis by evaluating its geometric consistency with the TSDF, whose construction is described later in the section. Although the hypotheses are generated from sparse tracked keypoints, the final selection uses dense 3D observations extracted from the current masked depth image. For each pose hypothesis, the dense observed points are transformed into the object frame, and the absolute TSDF value of the transformed points is computed. The hypothesis with the minimum TSDF score is selected.

\textbf{SDF Refinement.} 
To further improve the registration accuracy of the selected hypothesis $\bar{T}^{C}_{O}$, we perform an iterative pose refinement step that minimizes the dense point-to-implicit-surface distance induced by the TSDF. Given a dense point cloud $\mathcal{P}_{\mathrm{cur}}$ extracted from the current depth image, we formulate a robust nonlinear least-squares problem that drives the transformed observations onto the zero-level isosurface of the TSDF volume $\Phi(\cdot)$.

Let $T \in \mathrm{SE}(3)$ denote the object pose. Starting from the initial estimate $T_0=\bar{T}^{C}_{O}$, we iteratively solve for a pose increment $D \in \mathrm{SE}(3)$ by minimizing
\begin{equation}
D^{*} =
\underset{D \in \mathrm{SE}(3)}{\arg\min}
\sum_{p \in \mathcal{P}_{\mathrm{cur}}}
\rho_H\!\left(
\Phi\!\left(
T_i^{-1} D^{-1} p
\right)^2
\right),
\end{equation}
where $\rho_H(\cdot)$ denotes the Huber loss~\cite{huber1992robust}. After each iteration, the pose estimate is updated as $T_{i+1} = D^{*}T_i$ for $i=0,\ldots, L$.
The spatial gradients of the TSDF are computed numerically, and the optimal transformation $D^*$ is obtained using the Levenberg--Marquardt algorithm~\cite{levenberg1944method,marquardt1963algorithm}.

\subsection{Graph Optimization}
To maintain a globally consistent 3D map and correct for accumulated drift over time, for each object, we jointly optimize the keyframe poses and the keypoints using an online factor graph formulation. This optimization is performed whenever a new keyframe is created.

For factor-graph optimization, we adopt the equivalent perspective of a fixed object observed by a moving virtual camera, which yields a standard landmark observation model. As a result, the factor graph is parameterized using inverse pose variables $X_m \defeq T^{O}_C(m)$. Let $\mathcal{X} = \{X_0, \cdots, X_{M_i}\}$ be the set of inverse keyframe poses of a given object, and $\mathcal{P} = \{p_0, \cdots, p_{N_i}\}$ denote the 3D coordinates of the keypoints in the object frame. We solve for the optimal configurations for $\mathcal{X}$ and $\mathcal{P}$ by minimizing the following nonlinear least-squares objective function:
\begin{equation}
    \mathcal{L}(\mathcal{X},\mathcal{P}) = \mathcal{L}_{\mathrm{prior}} + \mathcal{L}_{\mathrm{reg}} + \mathcal{L}_{\mathrm{obs}}.
\end{equation}
For simplicity, we will drop the object index $i$.

\textbf{Prior Loss.} To prevent gauge ambiguity, the prior loss anchors the first keyframe pose to the identity:
\begin{equation}
    \mathcal{L}_{\mathrm{prior}} = \left\lVert \log(X_0)^\vee \right\rVert^2_{\Sigma_{\mathrm{prior}}},
\end{equation}
where $\log(\cdot)^\vee$ denotes the vectorized Lie algebra representation of the pose in $\mathfrak{se}(3)$ obtained via the logarithm map from $\mathrm{SE}(3)$, and $\|e\|^2_{\Sigma} = e^\top \Sigma^{-1} e$ denotes the Mahalanobis norm.

\textbf{Pose Consistency Loss.}
The pose consistency loss penalizes deviations of keyframe poses from the frame-to-map registration results. Let $\tilde{X}_{m}^{m-1}$ denote the estimated relative pose between two consecutive keyframes obtained from the registration. The pose consistency loss is defined using a relative motion constraint rather than an absolute pose estimate and is formulated as
\begin{equation}
\mathcal{L}_{\mathrm{reg}} =
\sum_{m=1}^{M}
\rho_H\!\left(
\left\lVert
\log\!\left(
(\tilde{X}_{m}^{m-1})^{-1}X_{m-1}^{-1}X_m
\right)^\vee
\right\rVert^2_{\Sigma_{\mathrm{reg},m}}
\right),
\end{equation}
where $\Sigma_{\mathrm{reg},m}$ is the registration covariance matrix, which may vary between keyframes. 
% \yh{Once I examined the objective function in (4), I better understood the role of this term. 
% However, I believe it represents more than simply ``temporal consistency and smoothness.'' When I first read that description, I expected something akin to a velocity regularization term, e.g.,
% $\left\| \log\!\left(X_{m-1}^{-1} X_m \right) \right\|^2,$
% which directly penalizes large inter-frame motion. Instead, the objective in (4) appears to enforce consistency with the front-end estimate,
% effectively encouraging $X_m$ to remain close to the pose predicted by the front-end registration.
% In this sense, it functions as a conservative refinement term: the optimization refines the estimate,
% but remains anchored to the front-end solution, which serves as the current best hypothesis.
% Conceptually, this term behaves more like a front-end pose prior or anchoring regularizer,
% rather than a temporal smoothness constraint. Renaming it accordingly (e.g., 
% ``front-end consistency term,'' ``pose prior,'' or ``front-end anchoring regularizer'')
% may improve clarity.}

\textbf{Observation Loss.}
Under the inverse pose parameterization, each keyframe pose $X_m$ acts as a virtual camera observing static keypoints $p_n$ defined in the object frame. 
Let $\Omega$ denote the set of keypoint observations, where $(m,n) \in \Omega$ indicates that keypoint $p_n$ is visible in keyframe $m$ with a valid depth observation.
Under this parameterization, $X_m^{-1}p_n$ represents the keypoint expressed in the camera frame of keyframe $m$. 
Let $\tilde{z}_{m,n} \in \mathbb{R}^3$ denote the corresponding 3D observation expressed in the camera frame. 
The observation loss is defined as the discrepancy between the predicted keypoint location $X_m^{-1}p_n$ and the observed 3D point $\tilde{z}_{m,n}$ for $(m,n) \in \Omega$.

To measure this discrepancy, we represent observations using a bearing–range representation (spherical coordinates) and define the loss as the sum of directional and radial errors. 
Let $\phi:\mathbb{R}^3 \rightarrow \mathbb{R}^4$ be a mapping that converts a 3D point $p$ to its bearing–range representation,
$\phi(p) = \left[\frac{p}{\|p\|},\, \|p\|\right]$,
where $\frac{p}{\|p\|}$ denotes the unit direction (bearing) and $\|p\|$ denotes the radial distance. The observation loss is then defined as 
\begin{equation}
\mathcal{L}_{\mathrm{obs}} =
\sum_{(m,n) \in \Omega}
\rho_H\!\left(
d^2\big(\phi(\tilde{z}_{m,n}), \phi(X_m^{-1}p_n)\big)
\right),
\end{equation}
where $d^2\big((u_1,r_1),(u_2,r_2)\big)$ denotes the summation of the angle between $u_1$ and $u_2$ and $|r_1-r_2|$. We refer the reader to the Appendix for detailed information.

% \begin{equation}
% \begin{bmatrix}
% \tilde{b} \\
% \tilde{r}
% \end{bmatrix}
% \ominus
% \begin{bmatrix}
% b \\
% r
% \end{bmatrix}
% \triangleq
% \begin{bmatrix}
% U(\tilde{b})^{\top} \mathrm{Log}_{S^2,\tilde{b}}(b) \\
% \tilde{r} - r
% \end{bmatrix},
% \end{equation}
% where $\mathrm{Log}_{S^2,\tilde b}(b) \in T_{\tilde b}S^2$ is the Riemannian logarithm map on the unit sphere $S^2$, and $U(\tilde b) \in \mathbb{R}^{3 \times 2}$ is an orthonormal basis of the tangent space $T_{\tilde b}S^2$.

% We then define the observation loss on the product manifold $\mathbb{S}^2 \times \mathbb{R}_+$ as
% \begin{equation}
% \mathcal{L}_{obs} = \sum_{(m,n) \in \Omega_m} \rho_H \left(\left\lVert \phi(\tilde{z}_{m,n}) \ominus \phi(X_m^{-1}p_n) \right\rVert^2_{\Sigma_{obs,mn}} \right),
% \end{equation}
% where we use the following local-coordinate subtraction on $\mathbb{S}^2 \times \mathbb{R}_+$ as:
% \begin{equation}
%     \begin{bmatrix}
%         \tilde{b}\\
%         \tilde{r}
%     \end{bmatrix} \ominus
%     \begin{bmatrix}
%         b\\
%         r
%     \end{bmatrix} \defeq
%     \begin{bmatrix}
%         U(\tilde{b})^T\mathrm{Log}_{\mathbb{S}^2,\tilde{b}}(b)\\
%         \tilde{r} - r
%     \end{bmatrix}
% \end{equation}
% where $\mathrm{Log}_{\mathbb{S}^2,\tilde b}(b)\in T_{\tilde b}\mathbb{S}^2$ is the Riemannian logarithm map on $\mathbb{S}^2$, and $U(\tilde b)\in\mathbb{R}^{3\times 2}$ is an orthonormal basis of $T_{\tilde b}\mathbb{S}^2$.

\textbf{Solving the Optimization.} 
We implement and solve this nonlinear least-squares problem using the Levenberg-Marquardt optimizer~\cite{levenberg1944method,marquardt1963algorithm} provided by the GTSAM library~\cite{gtsam,factor_graphs_for_robot_perception}. At each keyframe insertion, the factor graph is updated and re-optimized to obtain refined inverse keyframe poses $\mathcal{X}^*$ and keypoint coordinates $\mathcal{P}^*$. The optimized graph maintains a globally consistent keyframe-keypoint map, resulting in a more consistent pose tracking result.

\subsection{3D Reconstruction}
With the globally consistent object trajectory from the factor graph optimization, we aggregate the segmented RGB-D observations into a dense 3D model. We represent the target object using an object-centric volumetric grid defined in the object frame and fuse the depth measurements via a Truncated Signed Distance Function (TSDF)~\cite{curless1996volumetric}, where we follow the approach in~\cite{newcombe2011kinectfusion} for projective TSDF update. The TSDF is updated online whenever a new keyframe is added. For final mesh extraction, we use the marching cubes approach~\cite{lorensen1998marching} and apply a filter to remove the disconnected component introduced by depth noise. Beyond reconstruction, the fused TSDF also serves as a dense geometric representation for pose-hypothesis selection and refinement.

\section{Experiments}
We evaluate Point2Pose on two public RGB-D pose-tracking benchmarks and our newly introduced \texttt{YCBMultiTrack} dataset. We compare against BundleSDF~\cite{wen2023bundlesdf}, which performs tracking and reconstruction without object CAD models, and FoundationPose~\cite{wen2024foundationpose}, which is evaluated with provided object CAD models.

\subsection{Datasets}
\textbf{Public benchmarks.}
We evaluate on two standard RGB-D pose-tracking benchmarks: \hod~\cite{hampali2020honnotate} and \eoat~\cite{wen2020se}. Following BundleSDF~\cite{wen2023bundlesdf}, we use the \texttt{HO3D\_v3} evaluation split, which contains 13 hand-object interaction sequences with 4 objects. We also evaluate on 9 \eoat~sequences with 5 YCB objects manipulated by a dual-arm robot, where objects are generally smaller in the image, and ground-truth poses are manually annotated.

\textbf{YCBMultiTrack.}
% We evaluate our methodology using the YCB object dataset \cite{calli2015ycb}.
\multitrack is built from a subset of the YCB object~\cite{calli2015ycb} and contains both synthetic and real-world RGB-D sequences. The two subsets serve complementary evaluation purposes. The synthetic subset provides controlled multi-object tracking sequences with exact ground-truth poses from simulation, enabling systematic evaluation of multiple simultaneously tracked objects. The real-world subset focuses on more challenging interaction scenarios, including severe occlusion, complete object disappearance, and object re-entry, with ground-truth poses obtained from a motion-capture system. Together, these sequences complement existing object tracking benchmarks by covering both controlled multi-object evaluation and real-world occlusion-heavy tracking. Each sequence includes RGB-D images, SAM2~\cite{ravi2024sam} per-object segmentation masks, object canonical mesh poses in the camera frame $T^C_{M_i}(t)$ for each object $i$, camera intrinsics, and visibility annotations indicating whether each object is visible or fully occluded.

% Since $T^C_{O_{i,0}}=I$, we have $T^{O_i}_{M_i} = T^C_{M_{i,0}}$. Therefore, the ground truth pose of the object frame can be computed as $T^{C}_{O_{i,t}} = T^C_{M_{i,t}}(T^{O_i}_{M_i})^{-1}$.

% \yh{BELOW, we need to talk about (i) the number of video data, (ii) the number of objects in videos, (iii) some unique features in the dataset ... Synthetic data, gives us ground truth depths and pose labels. Real data, almost ground truth poses, depth from depth camera. masks are obtained by SAM.}

\begin{figure}[htbp]
    \centering
     \includegraphics[width=0.23\textwidth]{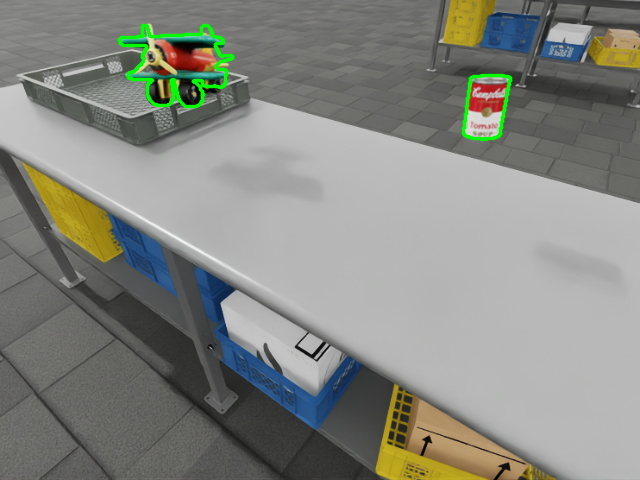}
     \includegraphics[width=0.23\textwidth]{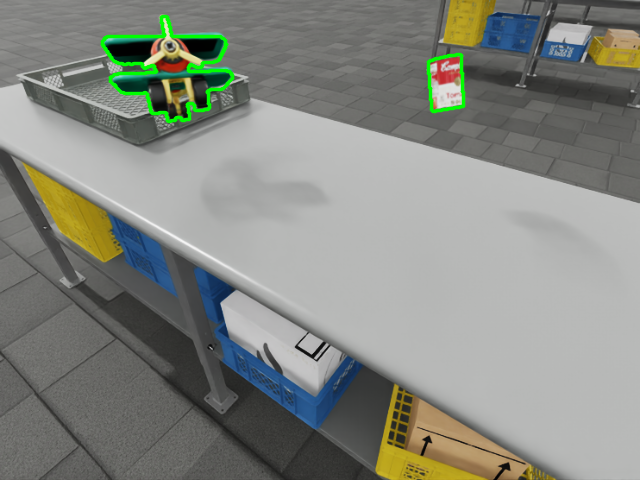}
     \includegraphics[width=0.23\textwidth]{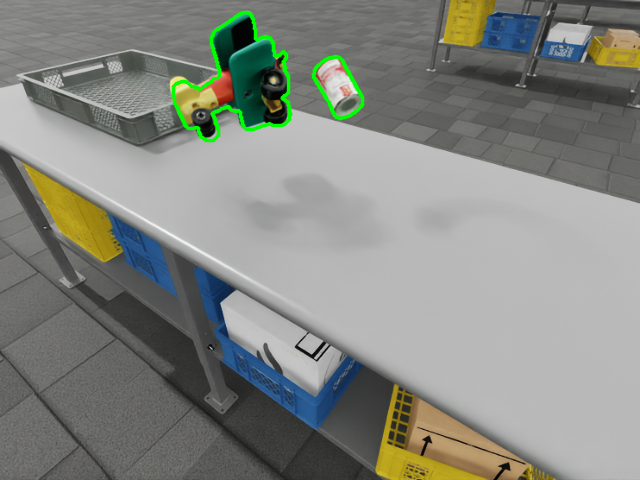}
     \includegraphics[width=0.23\textwidth]{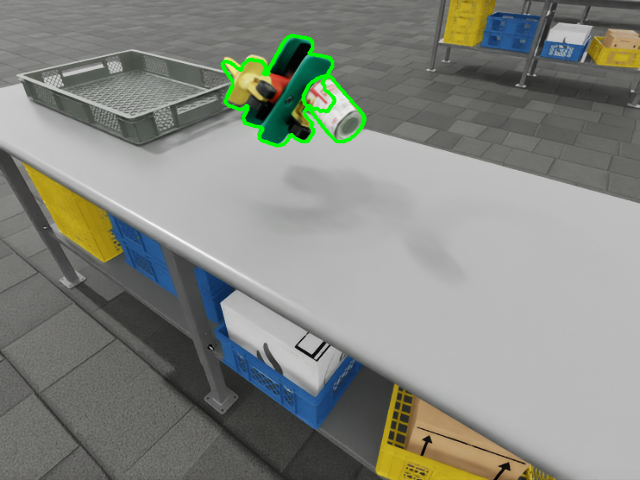}
    \caption{Example from the \texttt{YCBMultiTrack-Synthetic} dataset: a two-object sequence (toy airplane and tomato soup can) with circular motion and mutual occlusions.}
    \label{fig:synthetic_data}
\end{figure}

\textit{1. Synthetic.} % Due to a scarcity of open-source datasets providing comprehensive multimodal ground truth, specifically synchronized RGB sequences, camera-relative object poses, and segmentation masks, we developed a custom synthetic data generation pipeline. 
% Utilizing a simulation environment allows for the automated acquisition of precise ground truth labels that are often difficult or impossible to annotate manually in real-world settings. 
% We leverage Isaac Lab \cite{mittal2025isaac}, which supports GPU parallelization of physics simulations. 
% This framework allows us to generate environments with an arbitrary number of objects and complex trajectories in a parallelized manner.
% As demonstrated in Figure \ref{fig:synthetic_data}, we can efficiently simulate diverse scenarios ranging from single-object linear motion to multi-object configurations with complex occlusions. 
% This parallelized scheme enables the generation of high-quality, large-scale benchmark datasets in a matter of minutes, providing a scalable solution for training and evaluating object tracking and pose estimation tasks. 
We use seven objects from the YCB dataset and generate sequences containing one single-object and three two-object scenarios. Example data are shown in Fig.~\ref{fig:synthetic_data}. The objects move along either linear or circular trajectories to create diverse motion patterns and partial occlusion conditions. We simulate the motions and render RGB-D observations in Isaac Lab~\cite{mittal2025isaac}, which provides ground-truth depth and object poses. This allows us to evaluate tracking algorithms in a controlled setting with realistic object geometry and known ground truth.

% \footnote{Our synthetic data generation pipeline will be open-sourced, enabling users to efficiently generate custom multi-object tracking datasets in Isaac Lab using large-scale GPU-parallel simulation.}.

\textit{2. Real-World.} 
We use a total of five objects from the YCB dataset and generate sequences containing five single-object, four two-object, and two three-object scenarios. Example sequences are shown in Fig.~\ref{fig:real_data}.
We use an Intel RealSense D435i camera to collect RGB-D data. To obtain near ground-truth object pose labels, we use an OptiTrack motion capture system. Specifically, we attach four to five markers to each object and record their 6D poses during the sequence. The camera extrinsic matrix is calibrated using an AprilTag~\cite{wang2016apriltag} with known ground-truth pose.
After optimization-based time synchronization, we obtain aligned RGB-D frames and motion capture measurements for accurate pose labeling.

% \begin{figure}[htbp]
%     \centering
%     \includegraphics[width=\textwidth]{synthetic_dataset_cropped.pdf}
%     \caption{
%     % Visualization of the synthetic dataset.
%     % \textbf{(a) Left:} Single-object scenario (sugar box) following a linear trajectory.
%     % \textbf{(b) Center:} Two-object scenario (toy airplane and tomato soup can) following circular trajectories.
%     % \textbf{(c) Right:} Three-object scenario (potted meat can, pudding box, and mug) featuring a combination of linear and circular trajectories with potential occlusion behind the table. 
%     Overlayed images of a subset of the \textbf{YCBMultiTrack-Synthetic} dataset. \textit{Left}: single-object scenario (sugar box) following a linear trajectory. \textit{Center}: two-object scenario (toy airplane and tomato soup can) following circular trajectories with severe mutual occlusions. \textit{Right}: three-object scenario (potted meat can, pudding box, and mug) featuring a combination of linear and circular trajectories with occlusions caused by other objects and the table.
%     }
%     \label{fig:synthetic_data}
% \end{figure}

\textbf{Evaluation metrics.} 
Following prior work~\cite{wen2021bundletrack,wen2023bundlesdf,he2022fs6d}, we evaluate pose tracking with the area under the curve (AUC) of ADD and ADD-S, computed using the ground-truth mesh. ADD measures exact pose error and ADD-S accounts for object symmetries via closest-point matching. We evaluate reconstruction quality using Chamfer distance between the reconstructed and ground-truth meshes.

\subsection{Results on HO3D and YCBInEOAT}
% \Cref{tab:bundlesdf_point2pose_comparison} shows the results on the HO3D benchmark, where we compare our method with BundleSDF~\cite{wen2023bundlesdf}. Overall, Point2Pose achieves competitive performance across the evaluated sequences, with strong ADD-S AUC results and clear improvements on several challenging cases. The results on YCBInEOAT dataset are presented in Table~\ref{tab:ycb_ineoat}. Similar to HO3D, the proposed method achieves comparable performance to the state-of-the-art method. 
\Cref{tab:bundlesdf_point2pose_comparison} reports results on the \hod benchmark, and \Cref{tab:ycb_ineoat} reports results on the \eoat dataset. FoundationPose achieves strong pose tracking results, but it uses additional object-level information not assumed by BundleSDF or Point2Pose. Among methods without object CAD models, Point2Pose achieves comparable ADD-S AUC to BundleSDF on \texttt{HO3D}, with a higher mean ADD-S AUC, but lower ADD AUC and Chamfer Distance. On \texttt{YCBInEOAT}, BundleSDF achieves higher mean ADD-S and ADD AUC.

\begin{figure}[t]
    \centering
    \includegraphics[width=1.0\linewidth]{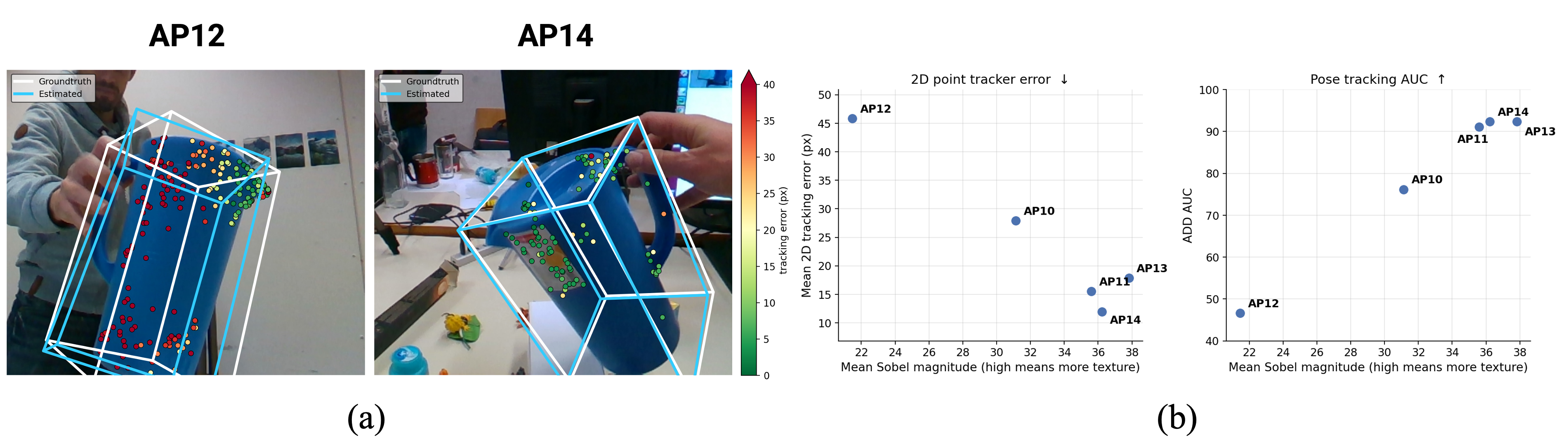}
    \caption{Texture analysis on \hod pitcher sequences. (a) Comparison of a low-texture view (\texttt{AP12}) and a view with stronger local texture (\texttt{AP14}). White and cyan boxes show ground-truth and estimated poses. In \texttt{AP12}, many high-error tracks occur on largely textureless surfaces. In contrast, the visible labels and cap in \texttt{AP14} provide stronger local texture, leading to more reliable low-error tracks, resulting in better pose alignment. (b) Mean Sobel gradient magnitude is used as a proxy for visible texture. Sequences with weaker texture generally exhibit larger 2D point-tracking errors and lower ADD AUC, suggesting that reduced correspondence reliability degrades downstream pose tracking.}
    \label{fig:results_failures}
\end{figure}

We observe lower ADD AUC on a subset of sequences, such as \texttt{AP12} in \hod and \texttt{bleach v1} in \eoat. These cases contain largely textureless visible object surfaces, making long-range point tracking more difficult. As shown in Fig.~\ref{fig:results_failures} (a), \texttt{AP12} exposes weaker texture than \texttt{AP14}, which contains clearer visual features such as white labels and a cap. This results in less stable 2D point tracks and reduced registration accuracy. A similar failure mode appears in the \eoat bleach sequences: in \texttt{v1}, only a small textured region is visible, which can induce geometric degeneracies in the tracked points. Nevertheless, Point2Pose still achieves reasonable ADD-S AUC in these edge cases, suggesting that the estimated poses often remain coarsely aligned even when orientation-sensitive ADD accuracy degrades.

To better understand the effect of object texture, we further analyze how visible texture influences point tracking and downstream pose tracking. We use Sobel gradient magnitude~\cite{gonzalez2018digital} as a quantitative proxy for local image texture. We evaluate 2D tracker accuracy by projecting object keypoints with ground-truth poses and comparing them with tracker predictions. Figure~\ref{fig:results_failures} (b) compares sequences containing the same pitcher object under different viewpoints, which expose different amounts of visible texture. We observe that lower-texture views generally lead to larger 2D point-tracking errors and lower pose-tracking accuracy. This suggests that low-texture views affect Point2Pose primarily by reducing the reliability of long-range 2D correspondences, which then degrades frame-to-map registration. Since our pipeline is modular with respect to the point tracker, future advances in texture-robust point tracking could directly improve this failure mode.

% \begin{figure}
%     \centering
%     \includegraphics[width=1.0\linewidth]{figures/new_failuires.png}
%     \caption{Comparison between low-texture and textured scenes. The green bounding box shows the pose estimated by the proposed method, while the red box denotes the ground truth. Point colors indicate the tracker’s uncertainty, with lighter colors corresponding to higher uncertainty and darker colors to more confident tracks; we visualize only points deemed visible by the tracker. In the \texttt{AP12} sequence, where our method attains a lower ADD score, many tracked points lie on largely featureless regions and exhibit high uncertainty. We additionally observe that these points fluctuate across frames, which leads to unstable correspondences and degrades pose estimation accuracy. In contrast, \texttt{AP14} provides richer texture, enabling more stable point tracks and more accurate pose estimates. A similar trend appears in the \texttt{Bleach} sequence: \texttt{v1} includes an extended interval dominated by a nearly uniform white side view, which yields unstable point tracks and consequently noisier pose estimates.}
%     \label{fig:results_failures}
% \end{figure}
\begin{table}[t]
\caption{Results on the \hod dataset~\cite{hampali2020honnotate}. ADD-S/ADD AUC are percentages over the $0$--$0.1\,\mathrm{m}$ threshold range. CD denotes Chamfer Distance for reconstruction evaluation. FoundationPose~\cite{wen2024foundationpose} is evaluated with provided CAD models. Bold numbers compare only between CAD-free methods.}
\label{tab:bundlesdf_point2pose_comparison}
\centering
\scriptsize
\setlength{\tabcolsep}{2.5pt}
\renewcommand{\arraystretch}{0.92}
\resizebox{\textwidth}{!}{%
\begin{tabular}{@{}lllccccc|ccccc|ccc|c@{}}
\toprule
Metric & Setting & Method
& \texttt{AP10} & \texttt{AP11} & \texttt{AP12} & \texttt{AP13} & \texttt{AP14}
& \texttt{MPM10} & \texttt{MPM11} & \texttt{MPM12} & \texttt{MPM13} & \texttt{MPM14}
& \texttt{SB11} & \texttt{SB13} & \texttt{SM1} & Mean \\
\midrule[0.1em]

\multirow{3}{*}{ADD-S (\%) $\uparrow$}
& CAD & FoundationPose
& 96.33 & 95.86 & 96.29 & 96.21 & 96.35
& 97.80 & 97.66 & 98.30 & 98.21 & 97.40
& 97.35 & 98.12 & Fail & 97.16 \\
\cmidrule(lr){2-17}
& \multirow{2}{*}{CAD-free} & BundleSDF
& \textbf{95.82} & 96.07 & \textbf{96.96} & 96.20 & \textbf{97.17}
& 90.94 & 96.29 & 96.16 & 59.05 & \textbf{97.18}
& \textbf{97.06} & \textbf{97.69} & \textbf{96.88} & 93.34 \\
& & Ours
& 93.71 & \textbf{96.18} & 84.33 & \textbf{96.27} & 96.13
& \textbf{95.31} & \textbf{97.08} & \textbf{97.12} & \textbf{96.58} & 96.42
& 94.35 & 97.58 & 89.15 & \textbf{94.63} \\

\midrule[0.1em]

\multirow{3}{*}{ADD (\%) $\uparrow$}
& CAD & FoundationPose
& 91.58 & 90.50 & 85.80 & 89.87 & 92.83
& 95.40 & 95.58 & 96.75 & 96.62 & 94.44
& 94.92 & 96.54 & Fail & 93.40 \\
\cmidrule(lr){2-17}
& \multirow{2}{*}{CAD-free} & BundleSDF
& \textbf{89.69} & 91.01 & \textbf{94.54} & \textbf{92.80} & \textbf{94.81}
& \textbf{78.78} & \textbf{91.69} & 91.33 & 33.71 & \textbf{94.16}
& \textbf{93.73} & \textbf{95.25} & \textbf{94.23} & \textbf{87.36} \\
& & Ours
& 76.19 & \textbf{91.12} & 46.64 & 92.36 & 92.39
& 73.99 & 88.16 & \textbf{93.86} & \textbf{63.19} & 85.56
& 85.99 & 94.90 & 65.95 & 80.79 \\

\midrule[0.1em]

\multirow{2}{*}{CD (cm) $\downarrow$}
& \multirow{2}{*}{CAD-free} & BundleSDF
& \textbf{0.52} & \textbf{0.58} & \textbf{0.66} & \textbf{0.66} & 0.98
& 0.71 & 0.49 & \textbf{0.48} & 0.71 & \textbf{0.47}
& \textbf{0.45} & \textbf{0.46} & \textbf{0.44} & \textbf{0.58} \\
& & Ours
& 1.20 & 0.89 & 2.97 & 0.96 & \textbf{0.86}
& \textbf{0.48} & \textbf{0.28} & 0.70 & \textbf{0.59} & 0.74
& 2.15 & 0.68 & 0.80 & 1.02 \\
\bottomrule
\end{tabular}
}
\end{table}

\begin{table}[t]
\centering
\caption{Results on the \eoat~dataset~\cite{wen2020se}. ADD-S/ADD AUC are percentages over the $0$--$0.1\,\mathrm{m}$ threshold range. FoundationPose~\cite{wen2024foundationpose} is evaluated with provided CAD models. Bold numbers compare only between CAD-free methods.}
\label{tab:ycb_ineoat}
\scriptsize 
\setlength{\tabcolsep}{3pt}
\renewcommand{\arraystretch}{0.92}

\resizebox{\textwidth}{!}{%
\begin{tabular}{@{}lllcccccccccc@{}}
\toprule
\multirow{2}{*}{Metric} & \multirow{2}{*}{Setting} & \multirow{2}{*}{Method} &
\multicolumn{2}{c}{\texttt{bleach}} &
\multicolumn{2}{c}{\texttt{cracker}} &
\multicolumn{2}{c}{\texttt{mustard}} &
\multicolumn{2}{c}{\texttt{sugar}} &
\texttt{tomato} &
\multirow{2}{*}{Mean} \\
\cmidrule(lr){4-5}\cmidrule(lr){6-7}\cmidrule(lr){8-9}\cmidrule(lr){10-11}\cmidrule(lr){12-12}
& & & v1 & v2 & v1 & v2 & v1 & v2 & v1 & v2 & & \\
\midrule[0.1em]

\multirow{3}{*}{ADD-S (\%) $\uparrow$}
& CAD & FoundationPose
& 96.78 & 95.59 & 96.41 & 93.30 & 97.70 & 97.00 & 97.10 & 93.79 & 96.35 & 96.00 \\
\cmidrule(lr){2-13}
& \multirow{2}{*}{CAD-free} & BundleSDF
& \textbf{94.00} & \textbf{95.47} & 95.60 & 91.72 & \textbf{96.32} & \textbf{96.28} & \textbf{96.12} & \textbf{90.56} & fail & \textbf{94.51} \\
& & Ours
& 82.88 & 93.90 & \textbf{96.37} & \textbf{92.50} & 95.26 & 95.65 & 94.31 & 87.60 & \textbf{95.52} & 92.67 \\

\midrule[0.1em]

\multirow{3}{*}{ADD (\%) $\uparrow$}
& CAD & FoundationPose
& 92.39 & 92.48 & 93.31 & 88.45 & 95.41 & 94.83 & 94.65 & 89.24 & 91.52 & 92.48 \\
\cmidrule(lr){2-13}
& \multirow{2}{*}{CAD-free} & BundleSDF
& \textbf{87.71} & \textbf{90.99} & 91.73 & 84.06 & \textbf{90.44} & \textbf{93.08} & \textbf{90.10} & \textbf{84.41} & fail & \textbf{89.07} \\
& & Ours
& 63.96 & 87.75 & \textbf{93.56} & \textbf{85.74} & 89.89 & 89.90 & 89.00 & 78.84 & \textbf{87.35} & 85.11 \\

\bottomrule
\end{tabular}%
}
\vspace{2pt}
% \footnotesize{BundleSDF fails on \texttt{tomato}; this case is excluded from its mean.}
\end{table}

% \begin{table}[t]
% \centering
% \caption{The results in the \eoat~dataset~\cite{wen2020se}. Comparison of BundleSDF~\cite{wen2023bundlesdf} and Ours by sequence group on ADD-S/ADD AUC. We refer the reader to the Appendix for the full sequence name.}
% \label{tab:ycb_ineoat}
% \scriptsize 
% \setlength{\tabcolsep}{3pt}
% \renewcommand{\arraystretch}{0.92}

% \resizebox{\textwidth}{!}{%
% \begin{tabular}{@{}clcccccccccc@{}}
% \toprule
% \multirow{2}{*}{Metric} & \multirow{2}{*}{Method} &
% \multicolumn{2}{c}{\texttt{bleach}} &
% \multicolumn{2}{c}{\texttt{cracker }} &
% \multicolumn{2}{c}{\texttt{mustard}} &
% \multicolumn{2}{c}{\texttt{sugar }} &
% \texttt{tomato} &
% \multirow{2}{*}{Mean} \\
% \cmidrule(lr){3-4}\cmidrule(lr){5-6}\cmidrule(lr){7-8}\cmidrule(lr){9-10}\cmidrule(lr){11-11}
% & & v1 & v2 & v1 & v2 & v1 & v2 & v1 & v2 & & \\
% \midrule

% \multirow{2}{*}{ADD-S (\%) $\uparrow$}
% & BundleSDF
% & \textbf{94.00} & \textbf{95.47} & 95.60 & 91.72 & \textbf{96.32} & \textbf{96.28} & \textbf{96.12} & \textbf{90.56} & fail & \textbf{94.51} \\
% & Ours
% & 82.88 & 93.90 & \textbf{96.37} & \textbf{92.50} & 95.26 & 95.65 & 94.31 & 87.60 & \textbf{95.52} & 92.67 \\
% \addlinespace[1pt]

% \midrule

% \multirow{2}{*}{ADD (\%) $\uparrow$}
% & BundleSDF
% & \textbf{87.71} & \textbf{90.99} & 91.73 & 84.06 & \textbf{90.44} & \textbf{93.08} & \textbf{90.10} & \textbf{84.41} & fail & \textbf{89.07} \\
% & Ours
% & 63.96 & 87.75 & \textbf{93.56} & \textbf{85.74} & 89.89 & 89.90 & 89.00 & 78.84 & \textbf{87.35} & 85.11 \\

% \bottomrule
% \end{tabular}%
% }
% \vspace{2pt}
% \end{table}

\begin{figure}[htbp]
    \centering
     \includegraphics[width=0.85\textwidth]{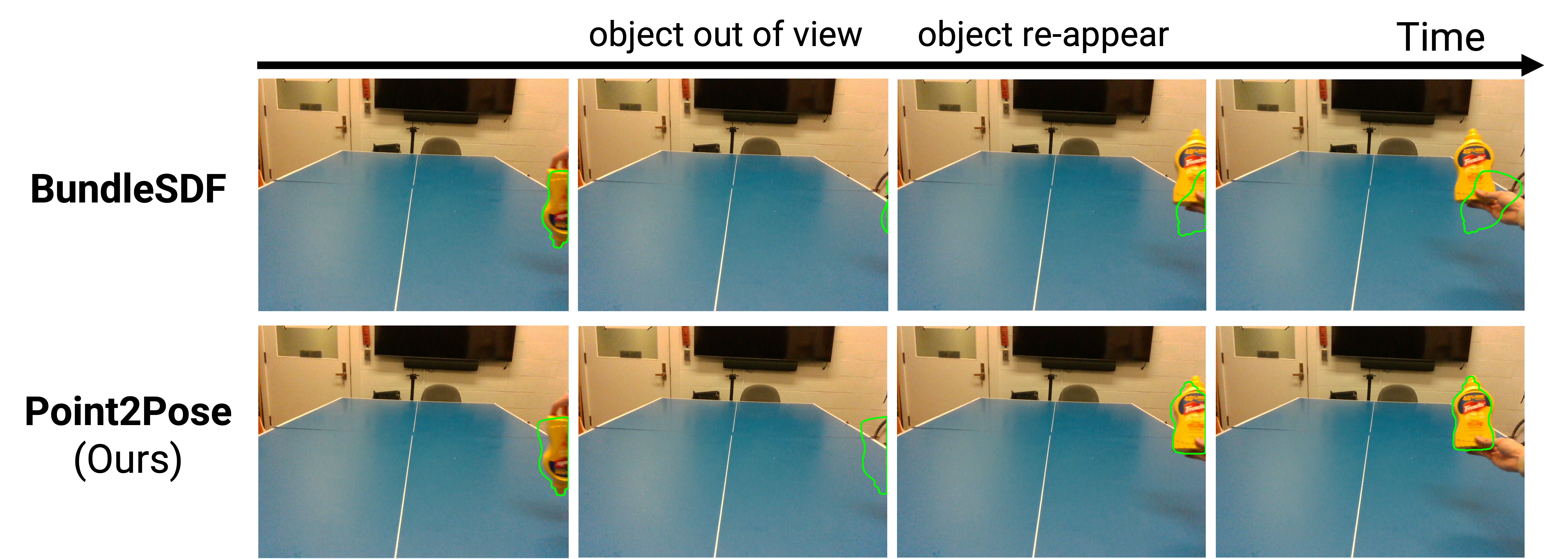}
    \caption{Results on a fully occluded scene from \texttt{YCBMultiTrack-Real}. After the object becomes completely occluded and later re-enters the scene, our method successfully recovers the object pose, whereas BundleSDF~\cite{wen2023bundlesdf} loses track and preserves a stale object orientation.}
    \label{fig:real_data_2}
\end{figure}

\subsection{Results on YCBMultiTrack}
We present results on the \multitrack dataset. Since both FoundationPose and BundleSDF support only single-object tracking, we run them sequentially for each object in a multi-object sequence. This setting gives the baselines a slight advantage in computational resources, as it never needs to share memory or compute across objects within a frame.

Table~\ref{tab:ycbmultitrack_syn} reports performance on the synthetic split, where accurate depth and ground-truth poses are available. Under these controlled conditions, FoundationPose achieves the best performance with the provided object meshes, as objects are largely continuously visible and long-term occlusion recovery is not strongly stressed. Among methods without object CAD models, our method performs consistently better across objects and scenes. Qualitative results of our method for the airplane–tomato sequence are shown in Fig.~\ref{fig:synthetic_data}. 

% Real-world results are shown in Table~\ref{tab:ycbmultitrack_real}. In these sequences, objects frequently undergo complete occlusion and may leave the camera view entirely, which substantially degrades BundleSDF and often causes failure. In contrast, our method remains reliable across sequences while tracking multiple objects simultaneously. Fig.~\ref{fig:real_data} visualizes a representative three-object scenario with severe arm-induced occlusions; when objects reappear, our method rapidly recovers their poses. Fig.~\ref{fig:real_data_2} shows a similar case in direct comparison with BundleSDF. 
% BundleSDF tracks correctly before the object exits the view, but upon re-entry its relocalization is biased by the previous estimate used as initialization, leading to an incorrect recovered pose. Our approach, instead, re-establishes correspondences and recovers the pose accurately after reappearance.
Real-world results are shown in Table~\ref{tab:ycbmultitrack_real}. In these sequences, objects frequently undergo complete occlusion and may even leave the camera view. Under these conditions, FoundationPose loses track despite using the provided object CAD mesh. BundleSDF exhibits a similar failure mode after complete occlusion. In contrast, our method remains reliable while tracking multiple objects simultaneously. Fig.~\ref{fig:real_data} shows a representative three-object scenario with severe arm-induced occlusions. When the objects reappear, our method quickly recovers their poses. Fig.~\ref{fig:real_data_2} presents a direct comparison with BundleSDF. While BundleSDF tracks correctly before the object leaves the view, its relocalization after re-entry is biased by the previous estimate used for initialization, resulting in an incorrect pose. These results highlight the key advantage of Point2Pose for simultaneous multi-object tracking and pose recovery after prolonged occlusion.

\begin{table}[t]
\centering
\caption{Results on the \multitrack dataset. ADD-S/ADD AUC are reported as percentages over the $0$--$0.1\,\mathrm{m}$ threshold range. FoundationPose~\cite{wen2024foundationpose}, denoted as FP, is evaluated with provided ground-truth object meshes, while BundleSDF~\cite{wen2023bundlesdf} and Ours do not use object CAD models. Bold numbers compare only between methods without object CAD models.}
\label{tab:object_metrics_syn_real}
\tiny
\setlength{\tabcolsep}{1.6pt}
\renewcommand{\arraystretch}{1.02}

\begin{subtable}[t]{0.49\textwidth}
\centering
\caption{YCBMultiTrack-Synthetic}
\resizebox{\linewidth}{!}{%
\begin{tabular}{@{}llccc|ccc@{}}
\toprule
\multirow{3}{*}{Category} & \multirow{3}{*}{Object} &
\multicolumn{3}{c|}{ADD-S (\%) $\uparrow$} &
\multicolumn{3}{c}{ADD (\%) $\uparrow$} \\
\cmidrule(lr){3-5}\cmidrule(lr){6-8}
& & CAD & \multicolumn{2}{c|}{CAD-free} & CAD & \multicolumn{2}{c}{CAD-free} \\
\cmidrule(lr){3-3}\cmidrule(lr){4-5}\cmidrule(lr){6-6}\cmidrule(lr){7-8}
& & FP & BundleSDF & Ours & FP & BundleSDF & Ours \\
\midrule[0.1em]

\multirow[t]{2}{*}{\makecell[l]{\\One\\object}}
& sugar   & 98.44 & 69.80 & \textbf{96.97} & 97.59 & 67.18 & \textbf{91.98} \\
& pitcher & 98.48 & \textbf{94.53} & 92.60 & 98.01 & 79.15 & \textbf{79.27} \\

\midrule

\multirow[t]{6}{*}{\makecell[l]{\\Two\\objects}}
& meat     & 98.57 & 76.67 & \textbf{97.13} & 98.03 & 72.13 & \textbf{93.06} \\
& sugar    & 99.07 & 72.09 & \textbf{96.87} & 98.84 & 69.16 & \textbf{92.31} \\
\cmidrule{2-8}
& scissors & 98.95 & 12.36 & \textbf{41.80} & 98.41 & 6.70 & \textbf{39.65} \\
& master   & 97.13 & 96.24 & \textbf{97.31} & 94.91 & 54.20 & \textbf{90.73} \\
\cmidrule{2-8}
& airplane & 99.01 & \textbf{96.02} & 88.46 & 98.60 & \textbf{92.21} & 55.07 \\
& tomato   & 97.43 & 96.17 & \textbf{98.20} & 96.21 & 63.66 & \textbf{81.41} \\

\midrule[0.1em]
& Mean & 98.39 & 76.74 & \textbf{88.67} & 97.58 & 63.05 & \textbf{77.94} \\
\bottomrule
\end{tabular}%
}
\label{tab:ycbmultitrack_syn}
\end{subtable}
\hfill
\begin{subtable}[t]{0.49\textwidth}
\centering
\caption{YCBMultiTrack-Real}
\resizebox{\linewidth}{!}{%
\begin{tabular}{@{}llccc|ccc@{}}
\toprule
\multirow{3}{*}{Category} & \multirow{3}{*}{Object} &
\multicolumn{3}{c|}{ADD-S (\%) $\uparrow$} &
\multicolumn{3}{c}{ADD (\%) $\uparrow$} \\
\cmidrule(lr){3-5}\cmidrule(lr){6-8}
& & CAD & \multicolumn{2}{c|}{CAD-free} & CAD & \multicolumn{2}{c}{CAD-free} \\
\cmidrule(lr){3-3}\cmidrule(lr){4-5}\cmidrule(lr){6-6}\cmidrule(lr){7-8}
& & FP & BundleSDF & Ours & FP
& BundleSDF & Ours \\
\midrule[0.1em]

\multirow[t]{5}{*}{\makecell[l]{\\One\\object}}
& tomato  & 45.10 & fail & \textbf{71.66} & 32.07 & fail & \textbf{36.60} \\
& mustard & 41.42 & 40.82 & \textbf{90.34} & 35.55 & 35.37 & \textbf{78.39} \\
& pudding & 13.86 & 44.48 & \textbf{90.81} & 4.21 & 24.87 & \textbf{73.22} \\
& meat    & 41.23 & 74.88 & \textbf{88.48} & 37.84 & 41.65 & \textbf{76.25} \\
& bleach  & 44.43 & 61.91 & \textbf{94.80} & 26.69 & 43.78 & \textbf{88.80} \\

\midrule

\multirow[t]{8}{*}{\makecell[l]{\\Two\\objects}}
& tomato  & 40.00 & 55.51 & \textbf{95.63} & 35.75 & 34.12 & \textbf{89.15} \\
& pudding & 42.04 & fail & \textbf{95.13} & 35.96 & fail & \textbf{91.63} \\
\cmidrule{2-8}
& tomato  & 33.71 & \textbf{93.48} & 87.21 & 31.75 & \textbf{83.44} & 44.87 \\
& pudding & 22.82 & fail & \textbf{84.99} & 13.09 & fail & \textbf{70.46} \\
\cmidrule{2-8}
& mustard & 18.22 & 12.19 & \textbf{95.35} & 12.11 & 12.04 & \textbf{91.02} \\
& meat    & 97.13 & fail & \textbf{96.49} & 94.48 & fail & \textbf{92.34} \\
\cmidrule{2-8}
& mustard & 29.01 & 70.08 & \textbf{91.01} & 27.08 & 29.19 & \textbf{81.72} \\
& meat    & 47.98 & fail & \textbf{90.02} & 40.82 & fail & \textbf{68.29} \\

\midrule

\multirow[t]{6}{*}{\makecell[l]{\\Three\\objects}}
& mustard & 18.18 & 72.90 & \textbf{85.88} & 17.60 & 39.04 & \textbf{74.16} \\
& meat    & 34.27 & fail & \textbf{92.21} & 32.90 & fail & \textbf{84.50} \\
& tomato  & 24.46 & fail & \textbf{90.36} & 19.96 & fail & \textbf{60.96} \\
\cmidrule{2-8}
& bleach  & 90.92 & fail & \textbf{92.65} & 83.72 & fail & \textbf{81.89} \\
& tomato  & 94.21 & \textbf{94.59} & 94.05 & 70.13 & 79.92 & \textbf{83.67} \\
& pudding & 28.23 & fail & \textbf{72.14} & 17.59 & fail & \textbf{41.75} \\

\midrule[0.1em]
& Mean & 42.49 & 62.08 & \textbf{89.43} & 35.23 & 42.34 & \textbf{74.17} \\
\bottomrule
\end{tabular}%
}
\label{tab:ycbmultitrack_real}
\end{subtable}

\vspace{2pt}
% \footnotesize{BundleSDF failed on several real-world sequences; failed cases are excluded from its mean.}
\end{table}

\subsection{Runtime}
Point2Pose runs at $2$--$10\, \mathrm{Hz}$ depending on tracker resolution and track count. The 2D point tracker is the main bottleneck, with runtime scaling roughly linearly with the number of tracks. Since the system is modular and our Python implementation is not runtime-optimized, faster point trackers or engineering improvements could further reduce latency.

\subsection{Ablation Studies}

\textbf{Component-Wise Ablation.}
We evaluate the contribution of the main modules on \hod in Table~\ref{tab:ablation_mean_std}. The no multi-hypothesis setting estimates the pose using a two-step procedure: an initial SVD-based alignment, followed by outlier removal using a fixed threshold and a second SVD using the remaining inliers. In the no SDF refinement setting, the pose from the multi-hypothesis stage is used directly without the SDF refinement step. In the no graph optimization setting, the estimated pose is used without performing graph-based global refinement. Removing the multi-hypothesis module leads to the largest performance degradation, while SDF refinement and graph optimization further improve pose accuracy through local and global refinement. Overall, these results confirm that the three modules are complementary and together enable robust and accurate pose estimation.

\begin{table}[t]
\centering
\caption{Component-wise ablation on \texttt{HO3D}.}
\label{tab:ablation_mean_std}
\tiny
% \scriptsize
\setlength{\tabcolsep}{5pt}
\renewcommand{\arraystretch}{0.9}
\begin{tabular}{lcc}
\toprule
Method & ADD-S (\%) $\uparrow$ & ADD (\%) $\uparrow$ \\
\midrule
Proposed & \textbf{95.07} & \textbf{82.76} \\
No multi-hypothesis & 89.81 & 65.52 \\
No SDF refinement& 94.40 & 77.37 \\
No graph optimization & 94.56 & 78.80 \\
\bottomrule
\end{tabular}
\vspace{-2mm}
\end{table}

\textbf{Long Range v.s. Short Horizon Tracking.}
To isolate the role of long-range tracking, we introduce P2P-SH($10$), where points are permanently discarded after $10$ consecutive invisible frames. All other components are unchanged. Across two occlusion-heavy \texttt{YCBMultiTrack-Real} sequences, Point2Pose recovers $5/5$ occlusion events within $30$ frames after re-emergence, whereas P2P-SH($10$) recovers $0/5$ events. On the mustard sequence, Point2Pose achieves $93.79$ ADD-S AUC, and $84.36$ ADD AUC, while P2P-SH($10$) degrades to $38.52$, and $31.44$, respectively. This ablation directly supports our claim that persistent long-range point identity is essential for recovery from prolonged or complete occlusion.

% \begin{table}[t]
% \centering
% \caption{Ablation results on keypoint sampling methods.}
% \label{tab:keypoint_sampling_mean_std}
% \small
% \begin{tabular}{lcc|cc}
% \toprule
% & \multicolumn{2}{c|}{ADD-S AUC (\%) $\uparrow$} & \multicolumn{2}{c}{ADD AUC (\%) $\uparrow$} \\
% \cmidrule(lr){2-3} \cmidrule(lr){4-5}
% Method & Mean & Std & Mean & Std \\
% \midrule
% Proposed & \textbf{95.07} & \textbf{3.74} & \textbf{82.76} & \textbf{15.18} \\
% Super Point Only & 93.09 & 4.14 & 77.61 & 19.78 \\
% Uniform Sampling & 91.47 & 5.92 & 72.31 & 17.89 \\
% \bottomrule
% \end{tabular}
% \end{table}

\section{Limitations}
% One limitation of the proposed method lies in the 2D point tracker: it relies on sufficiently discriminative image texture to maintain stable correspondences, and can become unreliable on textureless or repetitive surfaces. This property weakens the proposed method's ability to track low-texture objects. A second limitation is the dependence on accurate instance segmentation. When the segmentation mask is noisy or incomplete, the system may sample points on the background or neighboring objects, yielding incorrect correspondences that degrade registration and downstream optimization. A third limitation is that the approach can be memory-intensive for many objects, as the number of tracked points increases significantly. In practice, this can be mitigated through engineering choices (e.g., capping or subsampling points, pruning inactive keypoints) and by leveraging more efficient next-generation point tracking models. Lastly, our choice of a classical volumetric TSDF for reconstruction trades reconstruction fidelity for simplicity and runtime efficiency, and can be less accurate than recent learning-based reconstruction methods. An interesting direction for future work is to replace the TSDF module with a modern, efficient neural reconstructor (e.g., EfficientNeRF~\cite{hu2022efficientnerf}), while retaining online fusion and tight coupling with tracking.
The proposed method has several limitations. First, the underlying 2D point tracker relies on sufficiently discriminative image texture to maintain stable correspondences and may become less reliable on textureless or repetitive surfaces. 
Second, the approach depends on accurate instance segmentation; noisy or incomplete masks may introduce background points that degrade registration. 
Third, tracking many objects can increase memory usage due to the large number of tracked points, although this can be mitigated through practical engineering choices such as point subsampling or pruning inactive keypoints, as well as improved point-tracking models. 
Finally, our use of a classical TSDF representation prioritizes simplicity and runtime efficiency, but may yield lower reconstruction fidelity compared to recent learning-based approaches. 
Exploring modern neural reconstruction methods (e.g., EfficientNeRF~\cite{hu2022efficientnerf}) while maintaining online fusion and tight coupling with tracking is an interesting direction for future work.

\section{Conclusion}
We presented \textit{Point2Pose}, a causal, model-free method for multi-object 6D pose tracking and 3D reconstruction from monocular RGB-D video that can recover from complete occlusion. By leveraging modern 2D point trackers for long-range data association, our method maintains object identity through severe occlusions and enables recovery when objects reappear. We improve robustness under outlier-heavy correspondences with multi-hypothesis frame-to-map registration and TSDF-based hypothesis selection and refinement, while an online factor-graph optimization maintains global consistency and supports object-centric TSDF reconstruction. We also introduced \texttt{YCBMultiTrack}, a new synthetic and real-world benchmark for multi-object tracking under severe occlusions. Together, these results show that Point2Pose provides a practical model-free approach to multi-object tracking and reconstruction with recovery from complete occlusion.

% \clearpage  % TODO FINAL: This \clearpage needs to be removed from both review and camera-ready versions.

\section*{Acknowledgements}
This work was supported in part by the Technology Innovation Program (RS-2024-00427719) funded by MOTIE, Korea and the KIAT Global Industry Technology Cooperation Center Program (P246800183).

\appendix
\section{Implementation Details}
Below we describe the implementation details of the proposed method. Actual implementation, including code and parameters, will be available on the GitHub repository after the final decision. 

\subsection{2D Point Tracker}
We use the causal BootsTAP~\cite{doersch2023tapir,doersch2024bootstap} as the 2D point tracker to track sparse query points over time. Following~\cite{doersch2023tapir}, we use a visibility threshold of $0.5$. Each RGB frame is resized to $480\times480\times 3$ before being passed to the tracker, and the predicted point locations are mapped back to the original image resolution afterward.

\subsection{Point Sampling}
\textbf{Sampling Criteria.} Point sampling is triggered when the current tracked points no longer provide sufficient geometric coverage for reliable registration. In practice, we use two criteria: the estimated object rotation and the number of currently visible tracked points. We sample new points when the object rotates by more than \(10^\circ\) relative to previous sampled frames, or when the number of visible tracked points falls below $25$. 

\textbf{Point Promotion Strategy.} 
To avoid contaminating the object map with noisy depth measurements or short-lived tracks, newly sampled points are not trusted immediately. Once sampling is triggered, the new image points are added to the 2D tracker so that temporal correspondences can be established right away, but their associated 3D object-frame keypoints are initialized as tentative and are excluded from reliable frame-to-map registration. 

A pending point is promoted to a keypoint only if it passes a multi-frame verification process. Each pending point receives a score of +1 or +0 at every time step if it passes all of the following verificationchecks, and is promoted once this score reaches a consecutive streak threshold $N_{streak}$, which we set to $3$:

\begin{enumerate}
    \item Pose Stability: The relative pose change between consecutive frames must be small (e.g., rotation $< 2^\circ$ and translation $< 0.01$m) to ensure observations are captured during reliable tracking phases.
    \item Track Quality: The 2D track remains visible, has a valid depth value, and its tracking uncertainty falls below a strict threshold ($0.3$).
    \item Mask Consistency: The 2D projection of the track strictly lies within the current object segmentation mask.
\end{enumerate}
Once the consecutive streak threshold is reached, the point undergoes a final geometric verification before promotion. We evaluate the accumulated 3D observations of the pending point in the object coordinate frame to enforce spatial consistency. The point must satisfy a minimum observation count and exhibit a tight spatial spread, defined as a median absolute deviation of less than $0.008 m$. 

% \subsection{Frame-to-Map Registration for Pose Estimation}

\subsection{Graph Optimization}
\textbf{Observation Loss.}
We define the observation loss as:
\begin{equation}
\mathcal{L}_{\mathrm{obs}} =
\sum_{(m,n) \in \Omega}
\rho_H\!\left(
d^2\big(\phi(\tilde{z}_{m,n}), \phi(X_m^{-1}p_n)\big)
\right).
\end{equation}
We use the GTSAM~\cite{gtsam} implementation for the range and bearing loss, where the loss is defined over the product manifold $\mathbb{S}^2 \times \mathbb{R}_+$ as
\begin{equation}
d^2\big(\begin{bmatrix}\tilde{b} \\
\tilde{r}
\end{bmatrix}, \begin{bmatrix}
b \\
r
\end{bmatrix}) 
=
\begin{bmatrix}\tilde{b} \\
\tilde{r}
\end{bmatrix}
\ominus
\begin{bmatrix}
b \\
r
\end{bmatrix}
\triangleq
\begin{bmatrix}
U(\tilde{b})^{\top} \mathrm{Log}_{S^2,\tilde{b}}(b) \\
\tilde{r} - r
\end{bmatrix},
\end{equation}
where $\mathrm{Log}_{S^2,\tilde b}(b) \in T_{\tilde b}S^2$ is the Riemannian logarithm map on the unit sphere $S^2$, and $U(\tilde b) \in \mathbb{R}^{3 \times 2}$ is an orthonormal basis of the tangent space $T_{\tilde b}S^2$.

\subsection{3D Reconstruction}
We use the CUDA implementation of \texttt{tsdf-fusion-python} library~\cite{zeng20163dmatch} for 3D reconstruction. It follows the projective TSDF update approach. 

For a voxel center $\mathbf{v} \in \mathbb{R}^3$ in the object volume, its coordinates in the $m$-th camera frame are computed as $\mathbf{v}^m = T_0^{m*}\mathbf{v}$. We project this point onto the image plane using the camera projection function $\pi(\cdot)$ and sample the corresponding depth from the segmented depth image $\mathcal{D}_m$. 

We define the projective signed distance at voxel $\mathbf{v}$ as the difference between the measured depth and the voxel depth along the camera optical axis:
\begin{equation}
    d_m(\mathbf{v}) = \mathcal{D}_m(\pi(\mathbf{v}^m)) - [\mathbf{v}^m]_z,
\end{equation}
where $[\cdot]_z$ denotes the $z$-coordinate in the camera frame. To suppress noise and occlusion effects far from the observed surface, we use a truncation margin $\tau$. Following projective TSDF fusion, we discard voxels with $d_m(\mathbf{v}) < -\tau$. For valid voxels ($d_m(\mathbf{v}) \ge -\tau$), we define the normalized truncated TSDF observation as
\begin{equation}
    \Phi_m(\mathbf{v}) = \min\left(1, \frac{d_m(\mathbf{v})}{\tau}\right).
\end{equation}

The fused TSDF volume $V(\mathbf{v})$ and its corresponding weight volume $W(\mathbf{v})$ are updated by weighted averaging over valid observations:
\begin{equation}
    V(\mathbf{v}) = \frac{\sum_{m=0}^{M_i} w_m(\mathbf{v})\Phi_m(\mathbf{v})}{\sum_{m=0}^{M_i} w_m(\mathbf{v})},
\end{equation}
where $w_m(\mathbf{v})$ is set to zero if the projected voxel falls outside the image, outside the object mask, yields an invalid depth measurement, or is rejected by the truncation gate; otherwise, $w_m(\mathbf{v})=1$. RGB values are fused analogously from the corresponding color images to maintain a volumetric color map. To extract the dense 3D object mesh, one can use the Marching Cubes algorithm and compute the zero-crossing isosurface ($V(\mathbf{v})=0$) of the TSDF volume.

\section{Ablation Studies}
We conduct an additional ablation study on the HO3D dataset to analyze the effect of different keypoint sampling strategies. The results are 
summarized in Table~\ref{tab:keypoint_sampling_mean_std}.

Table~\ref{tab:keypoint_sampling_mean_std} shows that the proposed keypoint sampling strategy outperforms both SuperPoint-only and uniform sampling. 
Using only SuperPoint keypoints reduces performance, indicating that relying solely on detector-based keypoints may limit spatial coverage. Uniform sampling leads to an even larger drop, suggesting that randomly distributed points lack discriminative features for reliable matching. In contrast, the proposed strategy achieves the best accuracy and maintains relatively low variance, demonstrating that combining informative keypoints with better spatial coverage improves both robustness and pose estimation accuracy.

% \begin{table}[t]
% \centering
% \caption{Ablation results of the proposed method with different modules disabled. }
% \label{tab:ablation_mean_std}
% \small
% \begin{tabular}{lcc|cc}
% \toprule
% & \multicolumn{2}{c|}{ADD-S AUC (\%) $\uparrow$} & \multicolumn{2}{c}{ADD AUC (\%) $\uparrow$} \\
% \cmidrule(lr){2-3} \cmidrule(lr){4-5}
% Method & Mean & Std & Mean & Std \\
% \midrule
% Proposed & \textbf{95.07} & 3.74 & \textbf{82.76} & \textbf{15.18} \\
% No multi-hypothesis & 89.81 & 7.80 & 65.52 & 21.86 \\
% No SDF refine & 94.40 & 2.26 & 77.37 & 18.98 \\
% No graph optimization & 94.56 & \textbf{2.12} & 78.80 & 17.26 \\
% \bottomrule
% \end{tabular}
% \end{table}

\begin{table}[h!]
\centering
\caption{Ablation results on keypoint sampling methods.}
\label{tab:keypoint_sampling_mean_std}
\small
\begin{tabular}{lcc|cc}
\toprule
& \multicolumn{2}{c|}{ADD-S AUC (\%) $\uparrow$} & \multicolumn{2}{c}{ADD AUC (\%) $\uparrow$} \\
\cmidrule(lr){2-3} \cmidrule(lr){4-5}
Method & Mean & Std & Mean & Std \\
\midrule
Proposed & \textbf{95.07} & \textbf{3.74} & \textbf{82.76} & \textbf{15.18} \\
Super Point Only & 93.09 & 4.14 & 77.61 & 19.78 \\
Uniform Sampling & 91.47 & 5.92 & 72.31 & 17.89 \\
\bottomrule
\end{tabular}
\end{table}

% Please insert your acknowledgments here.

% ---- Bibliography ----
%
% BibTeX users should specify bibliography style 'splncs04'.
% References will then be sorted and formatted in the correct style.
%
\bibliographystyle{splncs04}
\bibliography{main}
\end{document}